\documentclass[preprint,11pt]{elsarticle}
\usepackage{epstopdf}
\usepackage{caption}
\usepackage{setspace}
\usepackage{amssymb}
\usepackage{multirow}
\usepackage{amsthm}
\usepackage{booktabs}
\usepackage{amsmath}
\usepackage{amssymb}
\usepackage{mathrsfs}
\usepackage{longtable}
\usepackage{lscape}
\usepackage{url}
\usepackage{tabularx}
\usepackage{epsfig}
\usepackage{colortbl}
\usepackage{subfigure}
\usepackage{tikz,mathpazo}
\usepackage{graphicx}
\usepackage{natbib}
\usepackage[noend]{algpseudocode}
\usepackage{algorithm}
\usepackage{algorithmicx}   
\usepackage{array}  
\usepackage{hyperref}

\usetikzlibrary{shapes.geometric, arrows}
\journal{XXXXXXX}

\begin{document}
	\begin{frontmatter}
		\title{Uncertainty Measurement of Basic Probability Assignment Integrity Based on Approximate Entropy in Evidence Theory}
		\author[address1]{Tianxiang Zhan}
		\author[address1]{Yuanpeng He}
		\author[address1]{Hanwen Li}
		\author[address1]{Fuyuan Xiao}
		
		\address[address1]{School of Computer and Information Science, Southwest University, Chongqing, 400715, China}
		\cortext[label1]{Corresponding author: Fuyuan Xiao, School of
			Computer and Information Science, Southwest University, Chongqing,
			400715, China. Email address: xiaofuyuan@swu.edu.cn,
			doctorxiaofy@hotmail.com.}
		\begin{abstract}
			Evidence theory is that the extension of probability can better deal with unknowns and inaccurate information. Uncertainty measurement plays a vital role in both evidence theory and probability theory. Approximate Entropy (ApEn) is proposed by Pincus to describe the irregularities of complex systems. The more irregular the time series, the greater the approximate entropy. The ApEn of the network represents the ability of a network to generate new nodes, or the possibility of undiscovered nodes. Through the association of network characteristics and basic probability assignment (BPA) , a measure of the uncertainty of BPA regarding completeness can be obtained. The main contribution of paper is to define the integrity of the basic probability assignment then the approximate entropy of the BPA is proposed to measure the uncertainty of the integrity of the BPA. The proposed method is based on the logical network structure to calculate the uncertainty of BPA in evidence theory. The uncertainty based on the proposed method represents the uncertainty of integrity of BPA and contributes to the identification of the credibility of BPA.
		\end{abstract}
		
	\end{frontmatter}
	
	\section{Introduction}
	Mass of information is generated every day. The information is not completely correct, which causes unnecessary trouble when making decisions. Evidence theory is a good way to resolve uncertainty \cite{dempster2008upper,rota1977mathematical}. As a method of uncertainty modeling, evidence theory has weaker conditions than Bayesian probability theory and has the ability to directly express unknown knowledge. Some information collected from the real world can be converted into basic probability assignment (BPA). Different BPAs can be obtained from different sources of information, and then a lot of potential information can be obtained by logical reasoning based on BPA. Evidence theory is widely used in expert systems, information fusion, intelligent decision-making and many other fields.
	
	Approximate entropy (ApEn) is a entropy used to quantify the amount of regularity and the unpredictability of fluctuations over time series data \cite{pincus1991approximate}. ApEn was developed by Steve M. Pincus by modifying an exact regularity statistic, Kolmogorov--Sinai entropy. ApEn is applied in finance, physiology, medicine science and so on. The ApEn of the network reflects the unpredictability of the network and can be used to measure the possibility of the existence of unknown nodes in the network \cite{west2012approximate}.
	
	Random walking refers to the inability to predict future development steps and directions based on past performance. The core concept is that the conserved quantity of any irregular walking person corresponds to a law of diffusion and transportation, which is close to Brownian motion, which is the ideal mathematical state of Brownian motion. At this stage, it is mainly used in Internet link analysis and financial stock market. In complex networks, random walks are often used for link prediction \cite{liu2010link}. Through the random walk in the complex network, the data of the starting node in the network can be distributed to the entire network, showing a kind of steady state.
	
	The existing BPA processing methods are to obtain a relatively reasonable BPA by combining multiple sets of BPA from different sources through some synthesis principles. But when the collected information is only a limited set of BPA, how to measure whether this BPA will achieve the desired effect or whether it is reasonable is an unresolved issue. This paper proposes the concept of BPA integrity, which can be used as a reference for the degree of trust in BPA. At the same time, this paper proposes a measure of the uncertainty of BPA integrity. Combining the process of BPA generation with random walks, a logical BPA network is proposed. Each node in the network corresponds to an element in the BPA. Through the network ApEn analysis of the possibility of the existence of potential nodes in the BPA network, the uncertainty of BPA integrity can be known.
	
	The structure of this article is as follows: The second section introduces the basic principles of network and ApEn. The third section explains how to calculate uncertainty. The fourth section specifically demonstrates how to calculate uncertainty. The fifth section will analyze the results of the fourth section and the uncertainty measure proposed. The sixth section summarizes the full text and mentions the future application of uncertainty measurement.
	
	\section{Preliminaries}
	\subsection{Basic probability assignment}
	Evidence theory suppoes the frame of discernment ($FOD$) which is the definition of a set of hypotheses as follows:
	$$\Theta =\left\{h_1,h_2,h_3,...,h_n\right\} \eqno(1)$$
	The power set $X$ of set $\Theta$ contains $2^n$ elements and defined as:
	$$X=\left\{\emptyset,\left\{h_1\right\},\left\{h_1,h_2\right\},\left\{h_1,h_2,h_3\right\},...,\Theta  \right\} \eqno(2)$$
	BPA is a mass function. And the element of $X$ must satisfy the properties as follows:
	$$\sum_{x\in X\left(\Theta \right)} m\left(x\right)=1 \eqno(3)$$
	$$m\left(x\right) \rightarrow \left[0,1\right] \eqno(4)$$
	\subsection{Approximate entropy}
	ApEn proposed by Pincus is a measure used to quantify the irregularity or complexity of a time series, and is a method of measuring the emergence of new patterns in a time series \cite{pincus1991approximate}.
	
	For a finite time series $u$ with $N$ time points, $u(i)$ represents the $i$-th time node in $N-m+1$. Divide $u$ into $N-m+1$ vectors $x(i)$ as follows:
	$$x\left(i\right)=\left[u\left(i\right),u\left(i+1\right),u\left(i+2\right),...,u\left(i+m-1\right)\right] \eqno(5)$$
	For two vectors $x(i)$ and $x(j)$, their distance $d[x(i),x(j)]$ is represented by the maximum value of the component difference as follows:
	$$d\left[x\left(i\right),x\left(j\right)\right]={max}_{k=1,2,...m}\left(\left|u\left(i+k-1\right)-u\left(j+m-1\right)\right|\right) \eqno(6)$$
	The ApEn of the time series u is calculated as follows:
	$$ApEn(u,r,N)=\Phi_m(r)-\Phi_{m+1}(r) \eqno(7)$$
	$$\Phi_m(r)=-\frac{1}{N-m+1}\sum_{i=1}^{N-m+1} ln\left(C_i^m\left(r\right)\right) \eqno(8)$$
	$$C_i^m\left(r\right)=\frac{\left|\left\{d[x(i),x(j)]|d[x(i),x(j)]<r,1\leqslant j \leqslant n-m+1\right\}\right|}{N-m+1} \eqno(9)$$
	\subsection{Approximate entropy for network}
	If $G$ is a network with N nodes then the slide approximate entropy of $G$ in dimension $m$ is
	$$SlideApEn(G,m) = ApEn(slide(G),m,r) \eqno(10)$$
	where $0<r<1$. The definition of function $slide(G)$ is as follows:
	$$slide(G)=(The\ sequence\ of\ all\ node\ degrees\ in\ the\ $$
	$$network\ G\ in\ descending\ order) \eqno(11)$$
	
	The ApEn of the network describes the possibility of undiscovered nodes in the network \cite{west2012approximate}.
	\section{Uncertainty measurement}
	\subsection{Integrity of basic probability assignment}
	We define the integrity of basic probability assignment. For a BPA, it is unknown whether the BPA is completely correct. But BPA may have the probability of inappropriate assignment. These probabilities should have been assigned to other elements, which did not appear in BPA or FOD. The integrity of BPA can be understood as whether the probability of BPA is correctly assigned, and there are no missing elements that are not assigned. When it is determined that the BPA is completely correct, the BPA is complete. There may be new elements in the BPA, indicating that completeness may be missing. When you don’t know whether BPA is completely correct, you need a possibility to measure the incompleteness of BPA.
	
	\subsection{Uncertainty measurement based on Approximate Entropy }
	In a known complex network, random walk is a method of link prediction that can discover potential structures in the network such as Fig.1. The random walk method first initializes the resource at a unique node. The ability of each side in the network to transmit resources is the same. After a sufficient number of wanderings, the resource assignment in the network is in a relatively stable state. 
	
	\begin{figure}[htbp]
		\centerline{\includegraphics[scale=0.25]{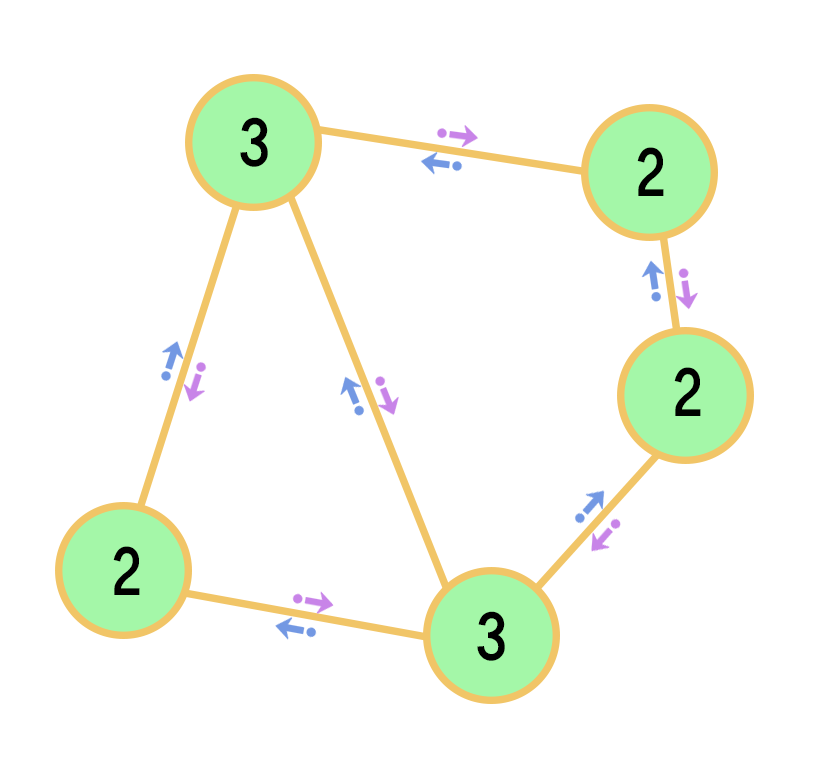}}
		\caption{Random walk schematic diagram (The number is the degree of the node)}
	\end{figure}
	
	For an actual BPA, the probability distribution of different elements is different. This BPA is in a stable state, and the sum of the probability distribution in the BPA is 1. Probability distribution in the BPA is equivalent to assigning the total probability 1 from the uncertain state (represented by the empty set  $\emptyset$ in the evidence theory) to other elements. Without knowing whether BPA is correct, the uncertainty of BPA needs to be measured. The same as the network is that a BPA assigns uncertain probabilities to known elements, and there is no correlation between known elements. Therefore, a BPA can be represented by a network, as shown in Fig.2. Except for the uncertain assignment in BPA, every other element is isolated. Therefore, the uncertain distribution node is related to the remaining elements, which are represented by edges in the network, and there is no edge disassociation between the remaining elements. The probability of uncertain assignment in BPA is represented by $m(\emptyset)$, so the probability that it can be assigned correctly is $1-m(\emptyset)$. In the network, it is equivalent to assigning probability 1 to the node that determines the element after a random walk. In the upper network, it is equivalent to the probability distribution 1 after a random walk, and the probability is assigned to the node that determines the element, which is a BPA. The number of edges between the uncertain distribution node and the remaining nodes is uncertain, and the ratio of the number of edges can be obtained by the probability of BPA.
	
	Because the remaining nodes are only linked to the uncertain assignment nodes, the relationship between the degree of the remaining nodes in the network, the number of edges between the remaining nodes and the uncertain assignment nodes and the degree of the remaining nodes is as follows:
	$$m(i)\sim Edge(i) \eqno(12)$$
	$$Edge(i)=Degree(i) \eqno(13)$$
	where $i$ represents the node of an element in the BPA. Therefore, the logical degree of the BPA network node is defined as follows:
	$$LogicalDegree(i)=m(i) \eqno(14)$$
	For the BPA network, the ApEn of the BPA network is calculated as follows:
	$$SlideApEn(G,m) = SlideApEn(LogicalGraph(BPA),m) \eqno(15)$$
	where $LogicalGraph(BPA)$ is BPA network and the degree of node is equal to logical degree to each elements.
	
	\begin{figure}[htbp]
		\centerline{\includegraphics[scale=0.3]{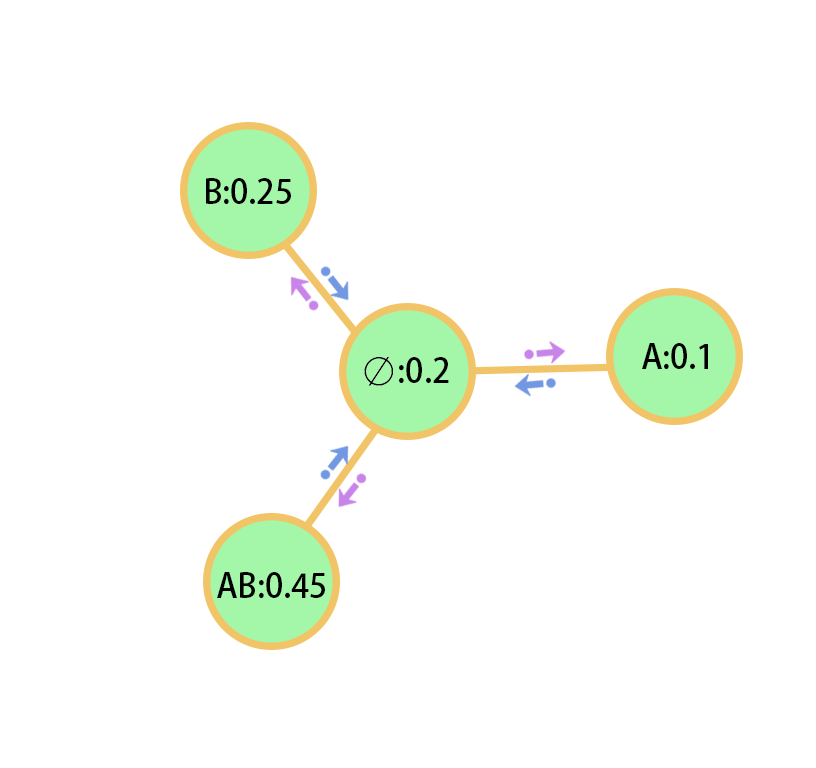}}
		\caption{Logical Graph of BPA $\left\{(\emptyset,0.2),(A,0.1),(B,0.2),(AB,0.45)\right\}$ }
	\end{figure}
	
	According to Pincus, when $ApEn(m,r,N)$ parameter is $m=2$ and $r=0.2*std(u)$, the value of $ApEn(m,r,N)$ has the least degree of dependence on $N$ where $N$ is the number of the time points in series $u$, resulting in a new model The more precise the description of the ability. So use the ApEn of the BPA network to measure uncertainty of BPA integrity (UI):
	$$UI(X)=SlideApEn(LogicalGraph(X),2) $$
	$$=ApEn(slide(LogicalGraph(X)),m=2 \eqno(16)$$
	$$,r=0.2*std(slide(LogicalGraph(X)))) \eqno(17)$$
	
	When BPA X has no probability assignment, or BPA X only assigns one element, the number of network nodes generated by X is less than 3, and the calculation cannot be completed. Because there is no probability assignment, and assigning only one element is meaningless. The BPA that assigns only one element indicates that the element is completely believed, and there is no possibility of undiscovered nodes. So it shows that the incomplete BPA only happena when the number of hypotheses identified by the frame of discernment is greater than 1.
	
	By measuring the possibility of undiscovered nodes in the BPA network, the possibility of incomplete BPA can be measured. The node of the BPA network is an element in the BPA. If a new element appears, it will cause an unreasonable distribution of the original BPA, which is an incomplete BPA. 
	
	\section{Experiment}
	
	In this section, a BPA example will be used to illustrate the operational process of uncertainty measurement.
	\begin{figure}[htbp]
		\centerline{\includegraphics[scale=0.3]{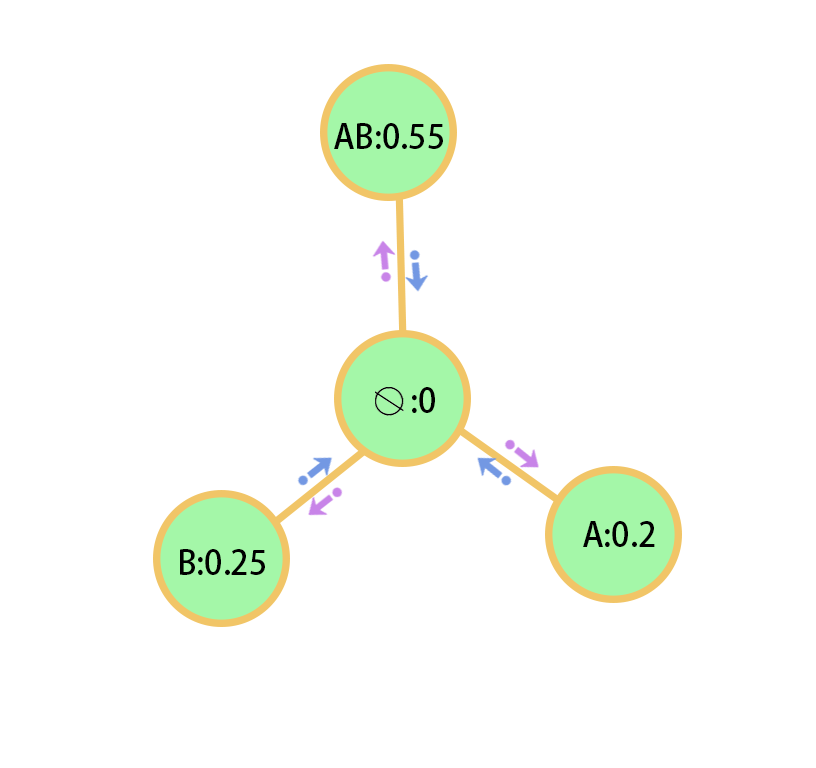}}
		\caption{Logical Graph of BPA X}
	\end{figure}	
	Now assume that there is a BPA as follows:
	$$X=\left\{(A,0.2),(B,0.25),(C,0.55)\right\} \eqno(18)$$
	The obtained BPA network is shown in Fig.3. At this time, the uncertain probability is assigned to 0. So the logical degree and slide degree sequence of X is as follows:
	$$LogicalDegree(X)=\left\{(A,0.2),(B,0.25),(AB,0.55),(\emptyset,0)\right\} \eqno(19)$$
	$$Slide(LogicalGraph(X))=(0.55,0.25,0.2,0) \eqno(20)$$
	Therefore, the BPA integrity uncertainty is as follows:
	$$UI(X)=ApEn((0.55,0.25,0.2,0),0.2,0.2*std((0.55,0.25,0.2,0))) \eqno(21)$$
	The uncertainty of the integrity $UI(X)$ of BPA X is equal to 0.4054.
	
	In addition, the uncertainty of the two sets of BPA $X_1$ and $X_2$ is additionally calculated. The definitions of $X_1$ and $X_2$ are as follows:
	$$X_1=\left\{(A,0.33),(B,0.33),(C,0.34)\right\} \eqno(22)$$
	$$X_2=\left\{(A,0.1),(B,0.1),(C,0.1),(\emptyset,0.7)\right\} \eqno(23)$$
	The logical degree and slide degree sequence of $X_1$ and $X_2$ is as follows:
	$$LogicalDegree(X_1)=\left\{(A,0.33),(B,0.33),(AB,0.34),(\emptyset,0)\right\} \eqno(24)$$
	$$Slide(LogicalGraph(X_1))=(0.33,0.33,0.34,0) \eqno(25)$$
	$$LogicalDegree(X_2)=\left\{(A,0.1),(B,0.1),(AB,0.1),(\emptyset,0.7)\right\} \eqno(26)$$
	$$Slide(LogicalGraph(X_2))=(0.7,0.1,0.1,0.1) \eqno(27)$$
	Therefore, the BPA integrity uncertainty is as follows:
	$$UI(X_1)=ApEn((0.33,0.33,0.34,0),0.2$$
	$$,0.2*std((0.33,0.33,0.34,0))) \eqno(28)$$
	$$UI(X_1)=0.05663301226513251 \eqno(29)$$
	$$UI(X_2)=ApEn((0.7,0.1,0.1,0.1),0.2,0.2*std((0.7,0.1,0.1,0.1))) \eqno(30)$$
	$$UI(X_2)=0.0566330122651324 \eqno(31)$$

	\section{Analysis}
	In the experimental part, the uncertainty $UI(X)$ of BPA $X$ is 0.4054. Assuming that there is a C hypothese in FOD that has not been identified, the definition of FOD is as follows:
	$$\Theta_{Actual} =\left\{A,B,C\right\} \eqno(32)$$
	$$X_{Actual}=\left\{\emptyset,\left\{A\right\},\left\{B\right\},\left\{C\right\},\left\{AB\right\},...,\Theta  \right\} \eqno(33)$$
	Assuming the real BPA X is as follows:
	\begin{table}[htbp]
		\centering
		\begin{tabular}{ccccccccc}
			\hline
			\textbf{Element}     & $\emptyset$ & A   & B   & C   & AB  & AC  & BC  & ABC \\ \hline
			\textbf{Probability} & 0           & 0.1 & 0.1 & 0.1 & 0.1 & 0.1 & 0.1 & 0.4 \\ \hline
		\end{tabular}
		\caption{The BPA of $X_{Actual}$}
	\end{table}
	Compared with the BPA of FOD without the C element, if the distribution of BPA is as follows:
	\begin{table}[htbp]
		\centering
		\begin{tabular}{ccccc}
			\hline
			\textbf{Element}     & $\emptyset$ & A   & B     & AB \\ \hline
			\textbf{Probability} & 0.7           & 0.1 & 0.1 & 0.1  \\ \hline
		\end{tabular}
		\caption{$X_{Actual}$ is transformed into a form of probability that does not know how to distribute it.}
	\end{table}
	It can be seen that incomplete FOD will have a huge impact on BPA, so it is important to measure the uncertainty of the integrity of BPA. Converting $X_{Actual}$ to a logical network, the calculated $UI(X_{Actual})$ is as follows:
	$$LogicalDegree(X_{Actual})=\{(A,0.1),(B,0.1),(C,0.1),(AB,0.1)$$
	$$,(AC,0.1),(BC,0.1),(ABC,0.4),(\emptyset,0)\} \eqno(34)$$
	$$slide(LogicalGraph(X_{Actual})=(0.4,0.1,0.1,0.1,0.1,0.1,0.1,0.1,0) \eqno(35)$$
	$$UI(X_{Actual})=0.0404 \eqno(36)$$
	
	Comparing the uncertainty of the actual BPA with the uncertainty of the original BPA, the uncertainty of the actual BPA is much smaller than the original BPA. It can be seen that when a BPA has relatively more existing elements, the uncertainty of the integrity of the BPA is also smaller. Considering the logical network of BPA, whenever FOD recognizes one more hypothesis, the maximum number of elements in BPA doubles, and the number of nodes doubles. Comparing these two BPA networks, when the number of nodes doubles, the possibility of undiscovered nodes will definitely become smaller, because the real BPA network is compared with the original BPA network, the real BPA network is The state after confirming the unknown node of the original BPA.
	
	For BPA with uncertain distribution probability, take $X_2$ as an example, and compare it with $X$.
	\begin{table}[htbp]
		\centering
		\begin{tabular}{ccccc}
			\hline
			\textbf{Element}     & $\emptyset$ & A   & B     & AB \\ \hline
			\textbf{Probability} & 0.7           & 0.1 & 0.1 & 0.1  \\ \hline
		\end{tabular}
		\caption{$X$}
	\end{table}
	\begin{table}[htbp]
		\centering
		\begin{tabular}{ccccc}
			\hline
			\textbf{Element}     & $\emptyset$ & A   & B     & AB \\ \hline
			\textbf{Probability} & 0           & 0.2 & 0.25 & 0.55  \\ \hline
		\end{tabular}
		\caption{$X_2$}
	\end{table}
	
	$X_2$ is actually the BPA converted by $X_{Actual}$ above. The uncertainty of $X_2,X_{Actual}$ is similar, and the uncertainty of $X$ is much greater than that of $X_2,X_{Actual}$. When the probability of existence in BPA does not know how to assign, it is equivalent to that the resources in the network are not completely transported out. When the remaining BPA does not know how to assign, the possibility of unknown nodes is small, and the network is not in a steady state. In a BPA network that is not in a steady state, it is equivalent to judging whether there are undiscovered nodes through the result of transporting fewer resources than the original. From the point of view of evidence theory, for BPA without uncertain probability, the probability of the sum of 1 needs to be correctly assigned to each element, while BPA with uncertain probability needs to assign the total probability to less than 1. When The probability of uncertainty in the BPA is clear, and the probability of incompleteness of this BPA decreases because the probability of assignment is reduced.
	
		\begin{figure}[htbp]
		\centering
		\subfigure
		{\includegraphics[width=10cm]{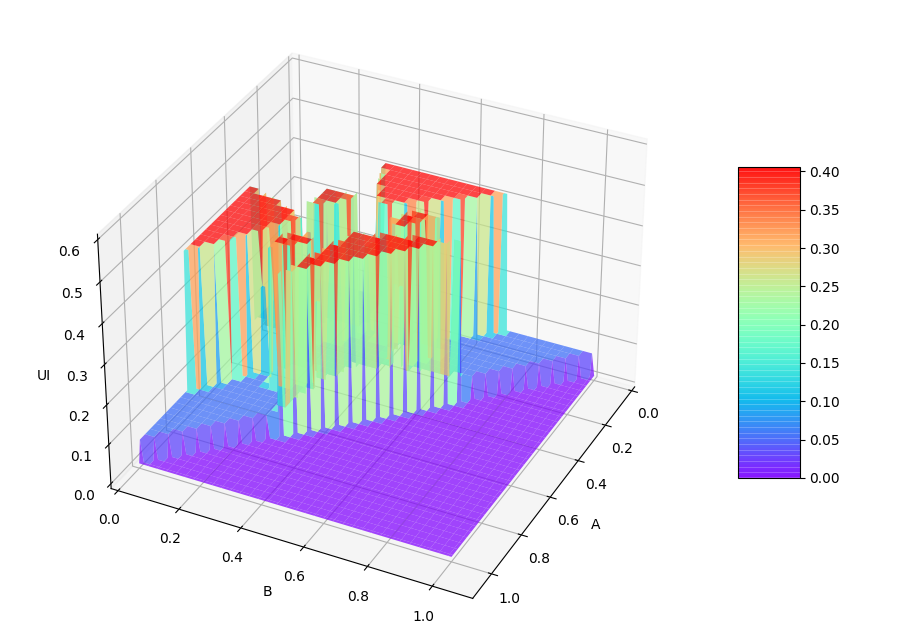}}
		\subfigure{\includegraphics[width=10cm]{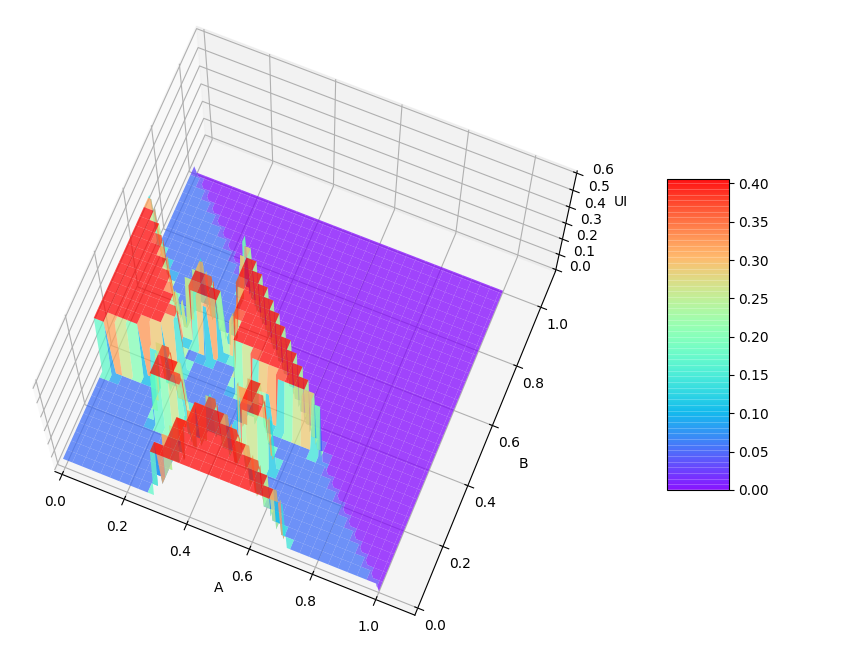}}
		\caption{UI of BPA $\left\{(\emptyset,0),(A,x),(B,y),(AB,1-x-y)\right\}$ } 
	\end{figure}
	
	The uncertainties of $X_Actual$ and $X_2$ are similar. On the one hand, $X_2$ is the BPA converted by $X_Actual$. On the other hand, the probability of not knowing how to assign is part of the probability of assigning to the correct element, which is reflected in $X_2$ by identifying an additional hypothesis C and adding 4 placement probabilities. More deeply, the uncertainty of $X_2$ is slightly higher than that of $X_Actual$, because if only $X_2$ can be determined, the specific situation of $X_Actual$ will not be known. Different situations of $X_Actual$ may translate into the same $X_2$. So the uncertainty of $X_2$ is higher.
	
	In a conclusion, the more elements in the BPA, the more the number of nodes in the BPA network, the lower the possibility of undetected nodes, and the lower the uncertainty of the integrity of the BPA.
	
	Here is a dynamic analysis of the simplest BPA. The definition of BPA $X_{test}$ is as follows:
	$$X_{test}=\left\{(A,x),(B,y),(AB,1-x-y),(\emptyset,0)\right\} \eqno(36)$$
	x and y are dynamically adjusted. For BPA, the value range of x and y is as follows:
	$$0\leqslant x \leqslant 1 \eqno(37)$$
	$$0\leqslant y \leqslant 1 \eqno(38)$$
	$$x+y \leqslant 1 \eqno(39)$$
	By determining the range of the $x$ and $y$ variables, here the uncertainty $UI(X_{test})$ of $X_{test}$ is used as the variable $z$, which is presented as a three-dimensional graph in the Fig.4.

	It can be seen from Fig.4 that $UI(X_{test})$ changes with $x$ and $y$ in layers, and there is no breakpoint on the surface formed by the entire UI. This is determined by the noise immunity and robustness of ApEn. The value of the $r$ parameter of the uncertainty UI in the calculation of ApEn is $0.2*std(X_{test})$. The value of $r$ eliminates the influence of the number of nodes on the uncertainty measurement to the greatest extent, and at the same time strengthens the anti-noise ability.

	\section{Conclusion}
	First and foremost, the integrity of the BPA is defined in this paper. Based on the network and ApEn characteristics, this paper proposes a measure of the uncertainty of BPA integrity. The uncertainty measurement of BPA supplements that a single BPA cannot use fusion methods to optimize BPA and cannot use existing information to know the reliability of BPA, so that a single BPA has a credibility evaluation standard. The proposed uncertainty measurement can be used as a weight or factor in BPA fusion or as a measure of uncertain systems in the future. The proposed uncertain measurement research will continue.

	\section*{Acknowledgment}
	%The authors greatly appreciate the reviews' suggestions and the editor's encouragement.
	This research is supported by the National Natural Science Foundation of China (No.62003280).

	\bibliographystyle{elsarticle-num}
	\bibliography{References}

\end{document}